\documentclass[10pt,twocolumn,letterpaper]{article}

\usepackage{ijcb}
\usepackage{times}
\usepackage{epsfig}
\usepackage{graphicx}
\usepackage{amsmath}
\usepackage{amssymb}
\usepackage{multicol}
\usepackage{multirow}
\usepackage{xcolor}
\usepackage{soul}
\usepackage{caption}

\newcommand*\samethanks[1][\value{footnote}]{\footnotemark[#1]}

\graphicspath{ {./images/} }

\ijcbfinalcopy 

\ifijcbfinal\fi
\begin{document}

\title{Landmark Enforcement and Style Manipulation for Generative Morphing}
\vspace{-5mm}
\author{Samuel Price\thanks{Authors Contributed Equally.}, Sobhan Soleymani\samethanks[1], Nasser M. Nasrabadi\\
West Virginia University\\
{\tt\small \{swp0001, ssoleyma\}@mix.wvu.edu, 
nasser.nasrabadi@mail.wvu.edu}
}

\maketitle
\thispagestyle{empty}

\begin{abstract}

    Morph images threaten Facial Recognition Systems (FRS) by presenting as multiple individuals, allowing an adversary to swap identities with another subject. Morph generation using generative adversarial networks (GANs) results in high-quality morphs unaffected by the spatial artifacts caused by landmark-based methods, but there is an apparent loss in identity with standard GAN-based morphing methods. In this paper, we propose a novel StyleGAN morph generation technique by introducing a landmark enforcement method to resolve this issue. Considering this method, we aim to enforce the landmarks of the morph image to represent the spatial average of the landmarks of the bona fide faces and subsequently the morph images to inherit the geometric identity of both bona fide faces. Exploration of the latent space of our model is conducted using Principal Component Analysis (PCA) to accentuate the effect of both the bona fide faces on the morphed latent representation and address the identity loss issue with latent domain averaging. Additionally, to improve high frequency reconstruction in the morphs, we study the train-ability of the noise input for the StyleGAN2 model. 
\end{abstract}

\section{Introduction}

\begin{figure}[t]
\begin{center}
\includegraphics[width=0.90\linewidth]{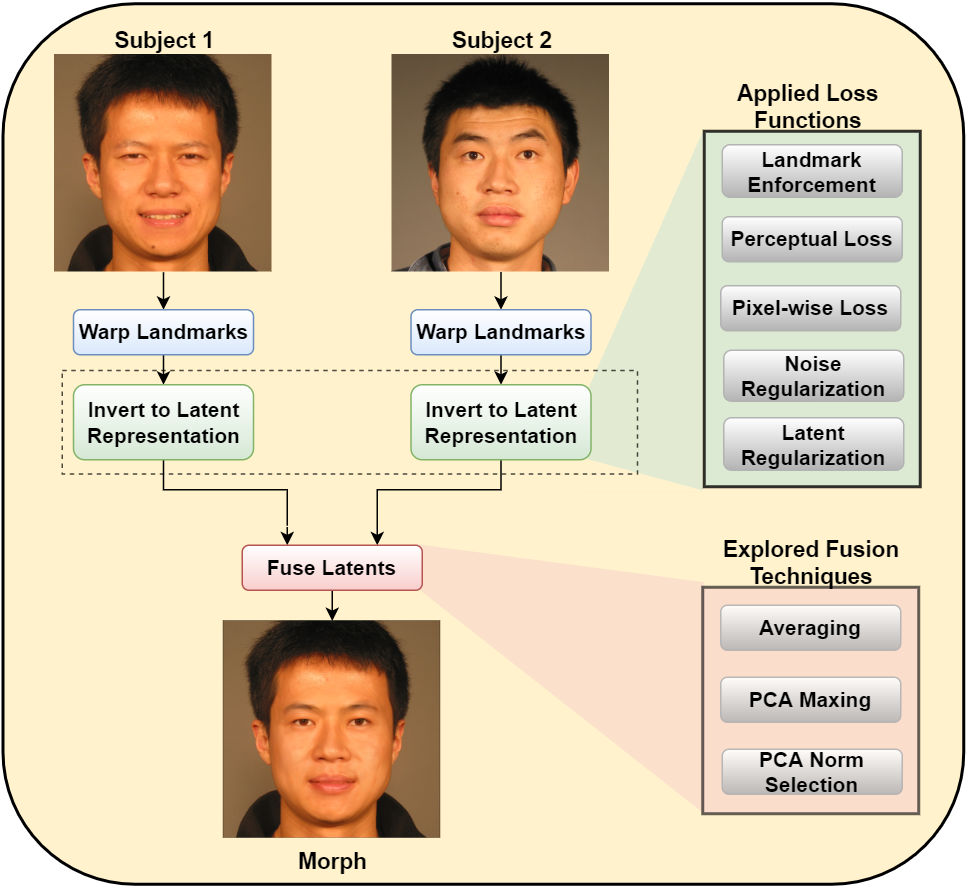}
\caption{Subjects are warped toward the average of their landmarks to produce a warped convex hull of each subject. The convex hulls are inverted into latent space of StyleGAN2 using a weighted combination of perceptual and pixel-wise losses in addition to latent and noise regularization exploring three techniques for blending latent codes.}
\label{fig::header}
\end{center}

\end{figure}

Generative Adversarial Networks (GANs) continue to grow in popularity in areas such as deepfake generation: realistic images generated by a deep neural network (DNN) \cite{goodfellow2014generative, korshunov2018deepfakes, wang2019towards}. With recent developments in the realistic face generation abilities of GANs \cite{karras2019style, karras2020analyzing}, the threat synthesized images pose to personal reputation, corporate sabotage, and national security grow concerning \cite{korshunov2018deepfakes}. As such, attacks on Facial Recognition Systems (FRS) mount as their usage continues to grow as an integral part of national security and law enforcement to verify identity \cite{bowyer2004face}. Border security is a key target as facial recognition is the only biometric required in electronic Machine-Readable Travel Documents (eMRTD) approved by the International Civil Aviation Commission \cite{ICAO}. Facial morph images have proven a threat to FRS when submitted by a bad actor to attack the enrollment stage of the biometric system integration guideline set by the ICAO, passing two safeguards: image tampering detection and identity verification \cite{ferrara2014magic}. A facial morph is an artificial face image generated by blending two or more bona fide face images of different individuals. The contributing subjects can use the morph for verification as FRS would find their identities indistinguishable to that of the morph. If a morph fools both the morph detector and is identified as the individual in question, a bad actor can circumvent these security measures. Using a GAN, our proposed technique generates morphs possessing the identity of two individuals to fool both human inspectors and FRS.

GAN-based morph generation blends the bona fide images in the latent space of the model by averaging the latent representations of contributing subjects \cite{damer2018morgan, venkatesh2020can_gan}. Improvements to early face generating GANs have increased their threat to FRS \cite{karras2019style, karras2020analyzing}. Although benefiting from enhanced visual quality, compared to other face morphing techniques, GAN-based face morphing falls short when used to attack FRS compared to landmark-based morphing due to a loss of identity in the morphed images \cite{quek2019face, venkatesh2020can_gan, zhang2021mipgan}. As presented in Figure \ref{fig::header}, we address this issue as we augment the latent space projections of the bona fide images by blending their landmarks before calculating their latent representations. Our landmark enforcement technique improves the morphed face's landmarks, being equidistant from the bona fide subjects' landmarks. To construct the latent representations for the bona fide subjects, we build upon inversion methods from \cite{abdal2019image2stylegan,karras2020analyzing, luxemburg2020stylegan2encoder} by incorporating a landmark enforcement algorithm to preserve the blended landmarks in the latent representation. In addition, we adapt the noise input of our model \cite{karras2020analyzing} to derive an improved image inversion algorithm resulting in latent codes with higher levels of reconstruction quality. 

We integrate our proposed inversion algorithm in the StyleGAN2 to improve the morph generation. We explore the constructed latent space using Principal Component Analysis (PCA) to enhance the blending of latent representations and further improve the quality of the morph images without adding additional optimization steps. This exploration aims at addressing the known issue with latent representation averaging which leads to morphs possessing biased or neither bona fide identities \cite{zhang2021mipgan}. We examine the covariance of latent representations using PCA and replace the latent code averaging with element-wise and vector-wise blending of PCA projected latent codes. By applying our image inversion algorithm and exploring latent representation blending in the PCA domain, we generate GAN-based morph images to fool FRS at increased rates while maintaining high image quality to fool a human inspector. Our major contributions in this paper are:

    \vspace{-1.5mm}

\begin{itemize}
    \item We present a novel StyleGAN2 morphing technique by enforcing landmarks to improve geometric identity preservation in the morph.
        \vspace{-0.25mm}

    \item We study latent space exploration in the PCA domain to improve latent code blending by addressing identity-imbalance issue.
            \vspace{-0.25mm}

    \item We study the influence of the noise input of our model to improve latent representations and morph image quality.
\end{itemize}

\section{Related Work}

\subsection{Landmark Morphing}

Facial morphing techniques split into two categories: landmark-based and GAN-based. GAN-based morphing operates in the latent space, whereas landmark-based morphing is performed in the image domain \cite{wu2011face, makrushin2017automatic, debruine2018webmorph, mallick2019opencv, quek2019face}. Landmark-based morphing uses landmark predictions of contributing subjects to warp them toward an equidistant set of landmarks. The pixel values of the warped images are alpha blended to complete the morph. Landmark-based morphing has been the most effective automated morphing threat to FRS \cite{venkatesh2020can_gan}; however, the blending of pixel values and imperfections in the landmark alignments create artifacts surrounding the morphed image eyes, mouth, nose, and edges around the face due to pasting. This ghosting effect increases the possibility of a human investigator recognizing the morph.

\subsection{GAN-Based Morphing}
Damer \etal \cite{damer2018morgan} introduced GAN-based morphing using MorGAN to invert images into latent representations via an encoder, averaging the faces in the latent domain, and inputting the resultant morph latent representation into the generator. In a study by Venkatesh \emph{et al.} \cite{venkatesh2020can_gan}, MorGAN morphs were shown to be limited in both image generation quality and output size of $64\times 64\times 3$. Morphs generated using MorGAN fail to pass the size standards set by the ICAO \cite{ICAO} while also failing to attack the verification of FRS.

\begin{figure*}[t]
\begin{center}
\includegraphics[width=\linewidth]{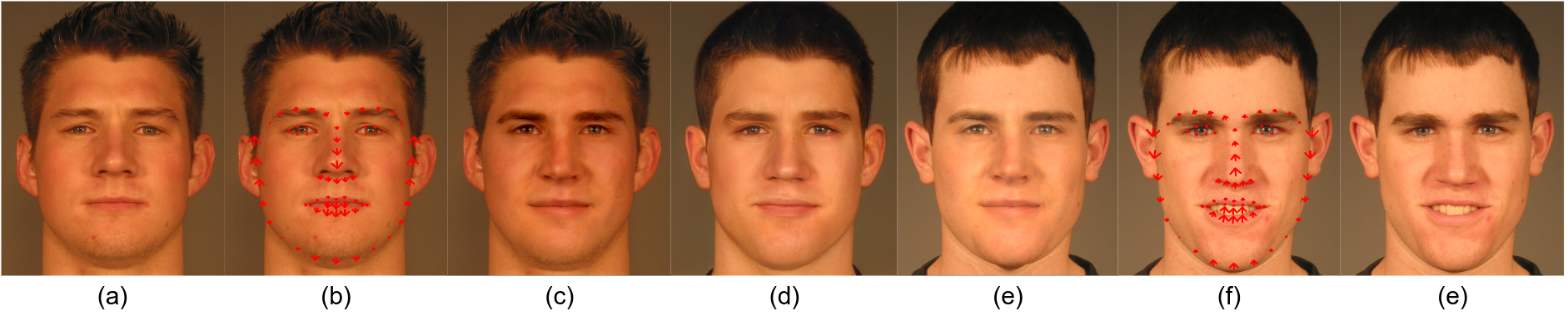}
\caption{Landmarks of bona fide images (a) and (e) are warped to an equidistant set of landmarks (b) and (f). Latent representations of the warped convex hulls are averaged, synthesized, and pasted on the background of the bona fide images to produce morphs (c) and (e). Without landmark warping, the morph image (d) generated using \cite{abdal2019image2stylegan} possess biased landmarks.}
\label{fig::landmark_banner}
\end{center}
\end{figure*}

Prior techniques for GAN-based morphing projects the average of two bona fide latent representations into the generator to synthesize the morphed image \cite{abdal2019image2stylegan,venkatesh2020can_gan}. The performance of morphs using StyleGAN \cite{karras2019style} significantly improved when compared to ones generated using MorGAN \cite{damer2018morgan}, but the performance is not comparable to landmark-based methods. Improvements to morph generation using StyleGAN include training encoders to estimate the latent embeddings \cite{richardson2020encoding, tov2021designing} or by adding new loss functions for optimization \cite{zhang2021mipgan}. MIPGAN \cite{zhang2021mipgan} proposed a hybrid approach to StyleGAN morphing by using both an encoder to estimate the latent codes of the bona fide subjects and an optimization cycle to improve the averaged latent code. The novel addition to their optimization cycle was an identity loss function using a pre-trained FRS model \cite{Deng2019arcface} to balance the identity of the morph between the bona fide subjects.

\subsection{PCA For Latent Exploration}

Principle Component Analysis (PCA) is widely used to evaluate correlation between samples in a dataset \cite{turk1991face}. The foundation of PCA calculates the covariance matrix of the dataset whose eigenvectors represent the variance of the entire dataset. Each eigenvector represents a variable amount of the variance of the dataset, so by removing the eigenvectors in order of smallest to greatest eigenvalue, the quality of the restored data degrades exponentially. PCA has been used to assist in the traversal and disentanglement of the latent space of GANs \cite{pham2020pcaae, yao2021latent}. We explore applications of PCA for blending two latent representations for morphing. Latent representation averaging has a limited success rate as the latent space is not a linear plane when using images not present in the latent space \cite{zhang2021mipgan}. By projecting the latent representations into the PCA domain, we explore the similarity of latent representations to improve the generated morphs.

\section{Methodology}

\begin{figure*}[t]
\begin{center}
\includegraphics[width=0.95\linewidth]{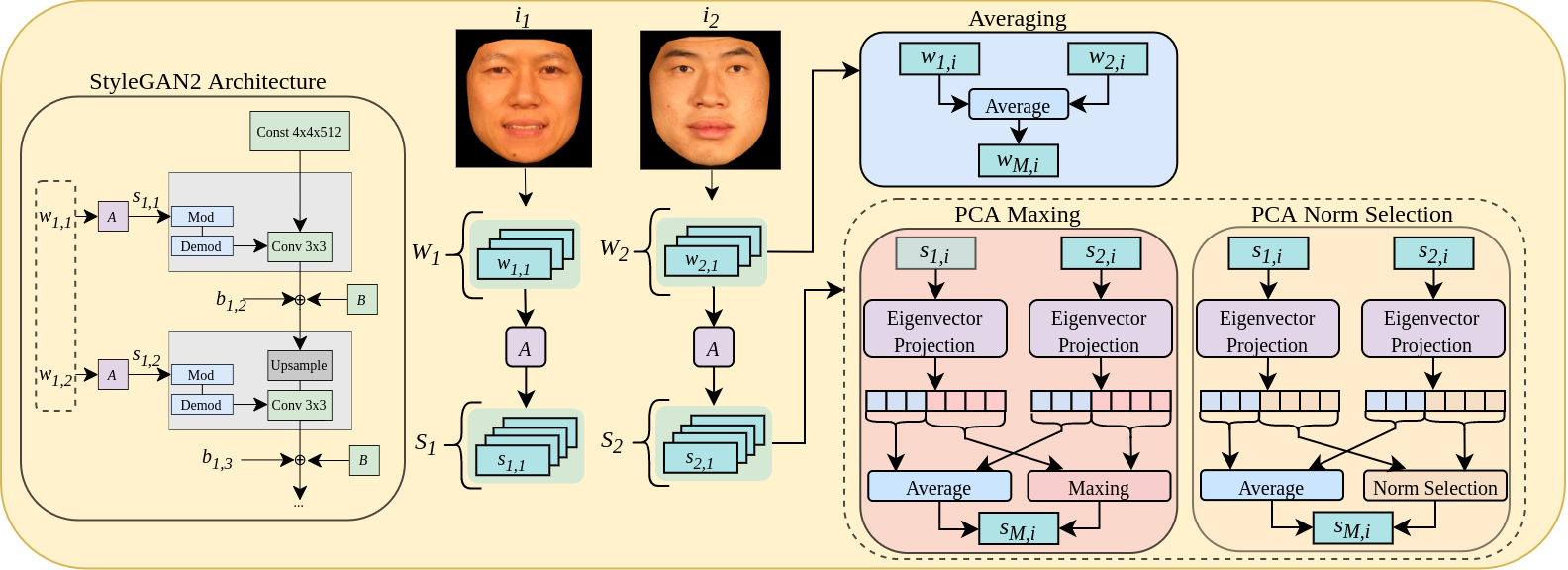}
\caption{Warped convex hulls of image $i_1$ and $i_2$ are inverted into their latent representations $W_1$ and $W_2$. The vectors are averaged to generate morph latent vector $w_{M,i}$. Latent codes $W_1$ and $W_2$ are put through the learned affine transform $A$ for each layer to produce style vectors $S_1$ and $S_2$. Style vectors are projected into the PCA model. The first projections are averaged and the remaining projections are blended using element-wise maxing or vector-wise norm selection to produce morphed style vector $s_{M,i}$. Noise input $B$ adds stochastic variation to the synthesized image.}
\label{fig::comprehensive}
\end{center}
\end{figure*}

Style-based generators \cite{karras2019style, karras2020analyzing} modify the latent input approach of \cite{goodfellow2014generative} to allow latent representations or styles to influence individual layers of the generator network directly. As presented in Figure~\ref{fig::comprehensive}, by progressively increasing the resolution of the convolutional layers, each layer achieves influence on different features of the output image. The early layers heavily influence the coarse features while the later layers influence finer details of the output image. By using a different latent code for each layer, an output image can be generated possessing a mixture of styles represented by the different latent codes, creating a new image. Morphing is an extension of style mixing as the styles of two images are blended to generate a morphed style.

We generate high-quality morph images utilizing a pre-trained StyleGAN2 model~\cite{karras2020analyzing}. To provide a better identity-preservation for the morphs, equidistant landmarks of the morph are enforced by warping the bona fide images' landmarks before latent optimization and preserving the warped landmarks through the addition of a landmark loss function. To remove potential artifacts caused when blending the exterior features of the original images (hair, ears, accessories), we embed convex hulls of our bona fide subjects. The projection of the morph's latent representation is pasted onto the original subjects' background, removing exterior artifacts. PCA decomposition of latent representations is explored to improve their blending for morph generation. The average representation can be biased toward one subject or possess neither identity \cite{zhang2021mipgan}. We use PCA to isolate the common variance (i.e., shared information) to average and blend the remaining using element-wise or vector-wise selection.

\subsection{Landmark Enforcement}
Each pair of facial images are first centered and cropped to $1024\times 1024\times 3$ due to StyleGAN2's difficulty in reconstructing images when faces are not properly centered \cite{kazemi2014millisecond}. Using Dlib~\cite{king2009dlib}, 68 landmarks are estimated for each bona fide subject~\cite{soleymani2021mutual}. The landmarks from a pair of subjects are averaged, generating an equidistant set of landmarks:
\vspace{-2mm}
\begin{equation}
    \label{landmark_warping}
    l_t(k) = \frac{1}{2}(M_k(i_1)+M_k(i_2)),\; \forall \; 1\leq k\leq n_l,
\end{equation}
where $M_k$ is the estimator for landmark $k$, $n_l$ is the total number of landmarks, $i_1$ and $i_2$ are the bona fide images, and $l_t$ is the equidistant set of target landmarks for the synthesized morphed image. 

We use Delaunay Triangulation to warp each bona fide subject's landmarks to the equidistant set \cite{quek2019face}. The pair of bona fide images now share a common set of landmarks. The artifacts caused by morphing latent representations of the hair, clothing, and accessories are removed by cropping out the face of the warped subjects. These convex hulls are generated by appending the boundary points of the face to the landmarks to create a mask \cite{opencv_library}. Applying the generated mask on the warped subjects, we isolate the warped faces from the pair of bona fide images.

To enforce the landmarks through the inversion process, we incorporate a landmark enforcement loss to preserve them in the latent representation. Through the optimization steps, the $L_2$ distance between the target's landmarks and the current synthesized image's landmarks is added to the total loss. The landmark enforcement loss is defined as:

\vspace{-2mm}
\begin{equation}
    \label{landmark}
    L_{land} = \sum_{k=1}^{n_l} (l_{t}(k) - M_k(g))^{2},
\end{equation}
where $l_t(k)$ is the target landmark $k$ and $M_k(g)$ is the synthesized image's landmark $k$. By warping the landmarks of the bona fide subjects and enforcing them when calculating latent representations, we produce geometrically unbiased morph images (see Figure \ref{fig::landmark_banner}). In addition, pasting the morphed masks onto the bona fide subjects' background greatly improves the visual quality of the morph images by blending the average pixel values with that of the bona fide background.

\subsection{Modified StyleGAN2 Inversion Method}

Morphing in the latent space requires inverting the bona fide subjects through the StyleGAN2 generator \cite{abdal2019image2stylegan, karras2020analyzing, luxemburg2020stylegan2encoder}. For perceptual quality assurance, we utilize the Learned Perceptual Image Patch Similarity (LPIPS) \cite{zhang2018unreasonable}. The LPIPS builds upon pre-trained convolutional neural networks (CNNs) \cite{simonyan2014very} to convert extracted features into an embedding for a given image. The target image \emph{t} and the synthesized image \emph{g} are first reduced to $256\times 256\times 3$ due to the input size of the feature extractor. We then take the cumulative squared distance between the embeddings of the target and the synthesized image to calculate the perceptual loss:
\vspace{-2mm}
\begin{equation}
    \label{lpips}
    L_{pert} =  ||E(t_d) - E(g_d)||_{2},
\end{equation}
where $E$ is the LPIPS embedding representation for the down-sampled images, $t_d$ is the down sampled target image, and $g_d$ is the synthesized image. Due to the down-sampling of the target and synthesized images for the perceptual loss, some information about the details in the image is lost. We add pixel-wise loss similar to \cite{abdal2019image2stylegan}, comparing the target image and synthesized image. We find that perceptual loss alone does not find the optimal embedding. The perceptual loss assists in finding the optimal region of the latent space whereas the pixel-wise loss improves the visual quality of the synthesized image as shown in Figure \ref{fig::loss_importance}. Pixel-wise loss is defined as:

\vspace{-2mm}

\begin{equation}
    \label{pixel-wise}
    L_{pix} = \frac{1}{N_{pix}}||t - g||_{1},
\end{equation}
where $N_{pix}$ is the size of the image.

The noise input ($B$ in Figure \ref{fig::comprehensive}) to the StyleGAN2 \cite{karras2020analyzing} generator is responsible for finer details or texture of the synthesized image \cite{karras2019style}. Optimization can be performed using a constant noise input generated before the optimization steps \cite{abdal2019image2stylegan} while only training for the optimal latent code. An alternative would be to train for a noise input along with the latent code. This leads to the noise incorporating too much information about the original subject, depreciating the quality of the latent code for morph generation. A noise regularization loss was introduced by Karras \emph{et al.} \cite{karras2020analyzing} to allow the noise to be trained along side the latent code while restraining the noise from learning structural information from the image. This regularization term forces each noise input to be a normally distributed signal:

\begin{figure}[t]
\includegraphics[width=0.95\linewidth]{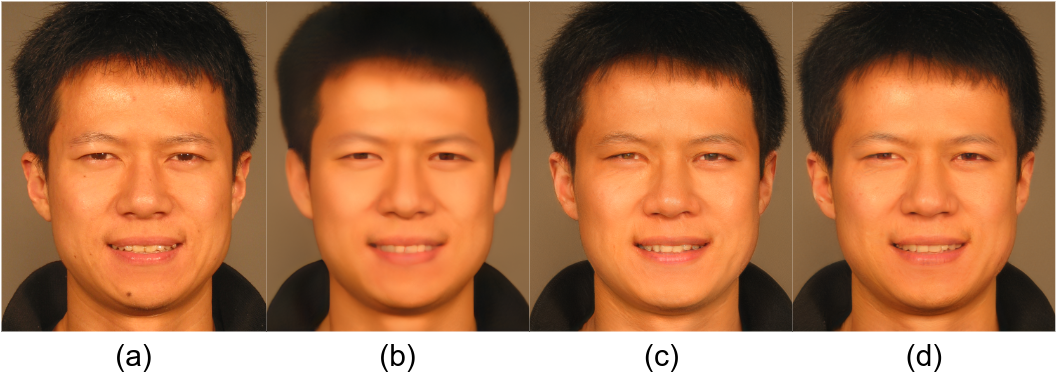}
\caption{(a) Bona fide image. Synthesized images using either pixel-wise loss (b) or perceptual loss (c) produces a non-optimal latent representation. Combining these losses (d) improves the overall quality of the synthesized image.}
\label{fig::loss_importance}
\end{figure}

\vspace{-3mm}

    \begin{equation}
    \label{noise_maps}
    \begin{split}
    L_{i,j} = ( 1/r_{i,j}^2 \sum_{x,y} n_{i,j}(x,y)n_{i,j}(x-1,y))^2\\
    + ( 1/r_{i,j}^2  \sum_{x,y} n_{i,j}(x,y)n_{i,j}(x,y-1))^2,
    \end{split}
    \end{equation}
where $n_{i,j}$ denotes noise map $i,j$ of noise input $B$, $r_{i,j}$ is the resolution of noise map $i,j$, $L_{i,j}$ is the regularization term for noise map $i,j$, and $(x,y)$ represents the spatial location. The noise regularization loss is defined as:
    \begin{equation}
    \label{noise_reg}
     L_{noise} = \sum_{i,j} L_{i,j}.
     \end{equation}

To prevent the latent code of each layer from going beyond the scope of the latent space, ultimately effecting the morph-ability of two subjects' latent codes, an $L_2$ penalty is applied to the latent codes \cite{luxemburg2020stylegan2encoder}. We weight the latent magnitude regularization penalty by a factor of $10^{-1}$, allowing for an accurate, but editable, latent representation to be found:

\begin{equation}
    \label{latent_reg}
    L_{lat} = \sqrt{\frac{1}{N_w} (W)^2},
\end{equation}
where $N_w$ is the size of latent code $W$ ($18\times 512 = 9216$). The total synthesis loss function is defined as:
\vspace{-2mm}
\begin{equation}
    \label{total_loss}
    L_{syn} = L_{pert} + \lambda_1L_{pix} + \lambda_2L_{noise} + \lambda_3L_{lat} + \lambda_4L_{land},
\end{equation}
where $\lambda_1$, $\lambda_2$, $\lambda_3$, and $\lambda_4$ are the scalar parameters for the individual losses.

\subsection{Influence of Noise}
\begin{figure}
\begin{center}
\includegraphics[width=0.95\linewidth]{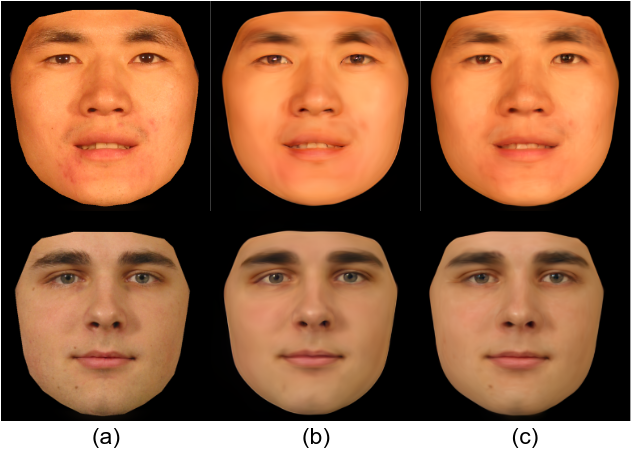} 
\end{center}
\caption{Bona fide images (a) compared to the synthesized images when training with noise (b) and without noise (c).}

\label{fig::noise_reg_images}
\end{figure}

Noise optimization is not the intended goal when optimizing for the latent space. The noise adds stochastic variation to the synthesized image, improving the visual quality of the image. Learning a complementary noise input while optimizing the latent code does assist in converging the loss early during training; however, our goal is finding the optimal latent code. Our work parallels that of \cite{abdal2020image2stylegan++} in that to find the optimal latent representation, the latent code and noise input must be trained separately. Removing noise from the optimization loop results in local minimas. We begin training for both the noise and latent representation until $T_s$ step at which point we remove the contents of each noise input, removing the noise's effect on the synthesized image. During the remaining steps, the latent representation for reconstructing the bona fide image is learned without the influence of noise, improving the reconstruction of the warped convex hulls (Figure \ref{fig::noise_reg_images}). Our modified inversion algorithm improves the learning of both coarser and finer details of the bona fide image. We note that without noise the synthesized images lack texture, increasing human detectability. After blending latent representations, we input noise generated for a random normal distribution when reconstructing the morph images to apply texture to the morph images. 

\subsection{Morphing in PCA Domain}

\begin{figure}[t]
\begin{center}
\includegraphics[width=\linewidth]{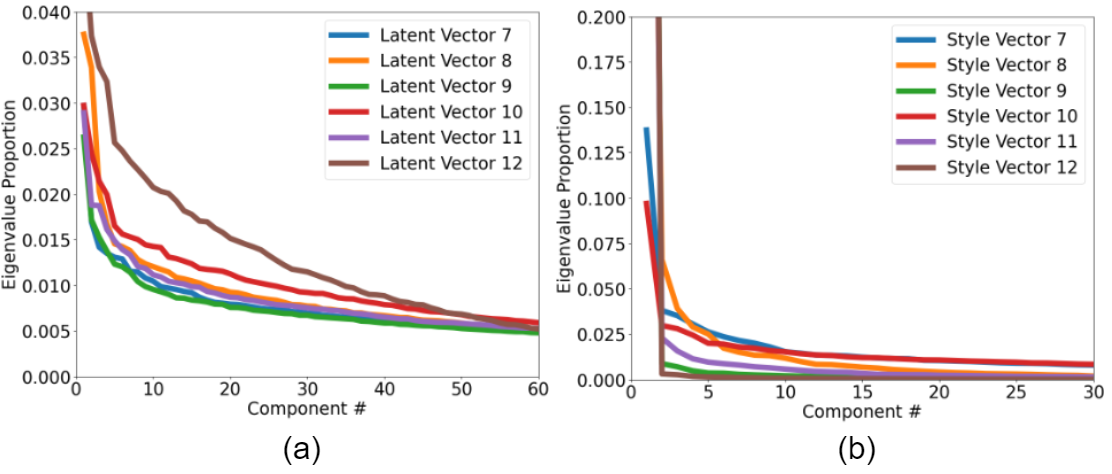}
\end{center}
\caption{Eigenvalue proportions for principal components of the sample latent vectors (a) and sample style vectors (b).}
\label{fig::pca_var}
\end{figure}

Morphs generated by averaging latent representations of bona fide subjects creates a high-quality image, but the identity of the morph is not guaranteed to be equidistant of the bona fide subjects \cite{zhang2021mipgan, sarka2022are}. This is primarily due to the bona fide images not existing in the constructed latent space of the model. We explore the identity imbalance problem of latent-based morphing using PCA. Latent codes are not directly inputted into the convolutional layers of the network. Learned affine transforms convert the latent vectors $w$ into true style vectors $s$ that influences the weights of the convolutional layer (see Figure \ref{fig::comprehensive}). This linear transformation changes the values and the dimensionality of the latent code to match the dimensionality of the layer. Unlike the latent codes of which we have 18, there are a total of 26 style vectors. This is due to the additional convolutional layers of the model used to convert the feature maps into an image \cite{karras2020analyzing}. The latent vector applied to the previous layer is inputted into the affine transform of these conversion layers, which generates addition style vectors. We project the transformed latent codes (styles) to the PCA domain. The eigenvalues calculated from the variance of the styles decay rapidly as shown in Figure \ref{fig::pca_var} in the right graph. With the improved eigenvectors, we explore ways to morph the style representations by only averaging projections on a first portion of the eigenvectors and varying the blending of the remaining eigenvectors, producing morphed style vectors $s_{M,i}$. Different amounts of variance are averaged to explore importance of high variance information in the latent representations.

As presented in Figure~\ref{fig::comprehensive}, we construct the PCA space using styles from embedded convex hulls. The total number of eigenvectors depends on the style vector we are projecting. Therefore, instead of a fixed number of eigenvectors used for averaging, we consider a percentage of eigenvectors $p$. We project each style vector into our pre-calculated PCA space, and consider the projection values using the first $p$ of eigenvectors. The first projections for a given morph pair are averaged to evenly blend the lower variance information of the two styles. For the projections from the remaining eigenvectors, we blend them using either the element-wise maxing or the $L_2$ norm vector-wise selection to blend the higher variance information.

For element-wise maxing, the goal is to mix the projected values without changing their value to address the identity loss when averaging the whole vector. For each style, blending is performed element-wise through the remaining projected values of a given pair to generate a new vector containing the maximum values between the two. We select the greater projected values between the two styles. The averaged and maxed vectors are added together, making the new morph style for the given pair. Our element-wise max blending is defined as:
\begin{figure}[t]
\begin{center}
\includegraphics[width=.95\linewidth]{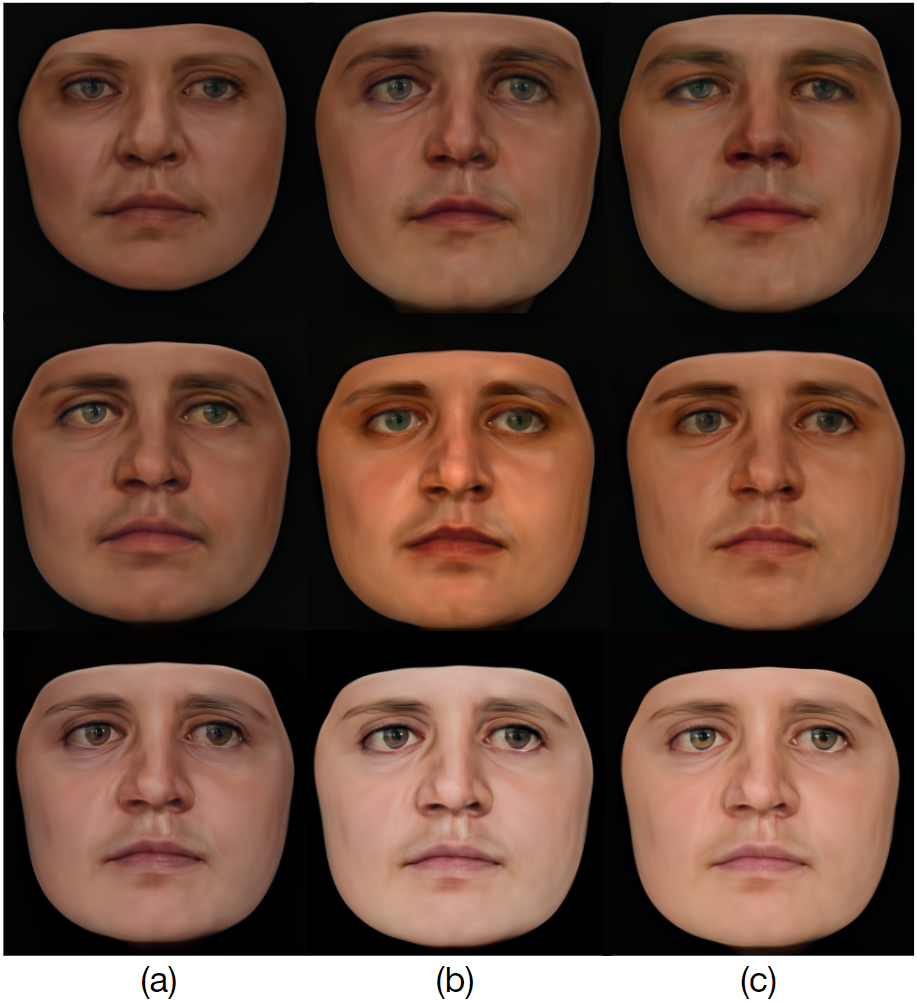}
\end{center}
\caption{Synthesized images after PCA projection using (a) 2\%, (b) 10\%, and (c) 20\% of the eigenvectors for coarse (top), intermediate (middle), and finer styles (bottom).}
\label{fig::pca_eigen}
\end{figure}

\vspace{-1mm}
\begin{equation}
    \label{pca_norm}
    \alpha_{M,i,j}=
    \begin{cases}
        \frac{1}{2}( \alpha_{1,i,j}+ \alpha_{2,i,j}) & \text{if $j \leq p e$}\\
        \max(\alpha_{1,i,j}, \alpha_{2,i,j}) & \text{else}
    \end{cases}
    ,
\end{equation}
where $\alpha_{1,i,j}$ and $\alpha_{2,i,j}$ are subject 1's and subject 2's $i^{th}$ style vector projection values onto the $j^{th}$ eigenvector of this style, $v_{i,j}$, respectively and $e$ is the total number of eigenvectors. Then, the reconstructed morphed style vector $s_{M,i}$ is given by: $s_{M,i}=\sum_j{\alpha_{M,i,j}v_{i,j}}$.

The $L_2$ norm selection technique uses a vector-wise selection as opposed to the element-wise selection. Where element-wise max aims to keep the original projected values, vector-wise norm keeps original projected style vectors. Keeping the entire projected vector preserves the projection of a known style vector. This removes error produced when traversing through the latent space by blending two different vectors. After computing the projection of the remaining eigenvectors, we compute the $L_2$ norm of the projected style vectors. We select the projected style vector with the largest $L_2$ norm and concatenate it to the averaged projection. Our vector-wise norm selection blending is defined as:

\vspace{-4mm}

\begin{multline}
\small
    \label{pca_max}
    \alpha_{M,i,j} = 
    \begin{cases}
        \frac{1}{2}( \alpha_{1,i,j}+ \alpha_{2,i,j}) & \text{if $j \leq p e$}\\
        \alpha_{1,i,j} & \hspace{-16mm}\text{else if $ ||P^*_i(s_{1,i})||_2>||P^*_i(s_{2,i})||_2$}\\
        \alpha_{2,i,j} & \text{else}
    \end{cases}
    ,
\end{multline}
\vspace{-3mm}
where $P^*_i(s_{1,i})=\sum_{j=pe+1}^{e}{\alpha_{M,i,j}v_{i,j}}$.

\section{Experiments}

\begin{figure}[t]
\begin{center}
\includegraphics[width=.95\linewidth]{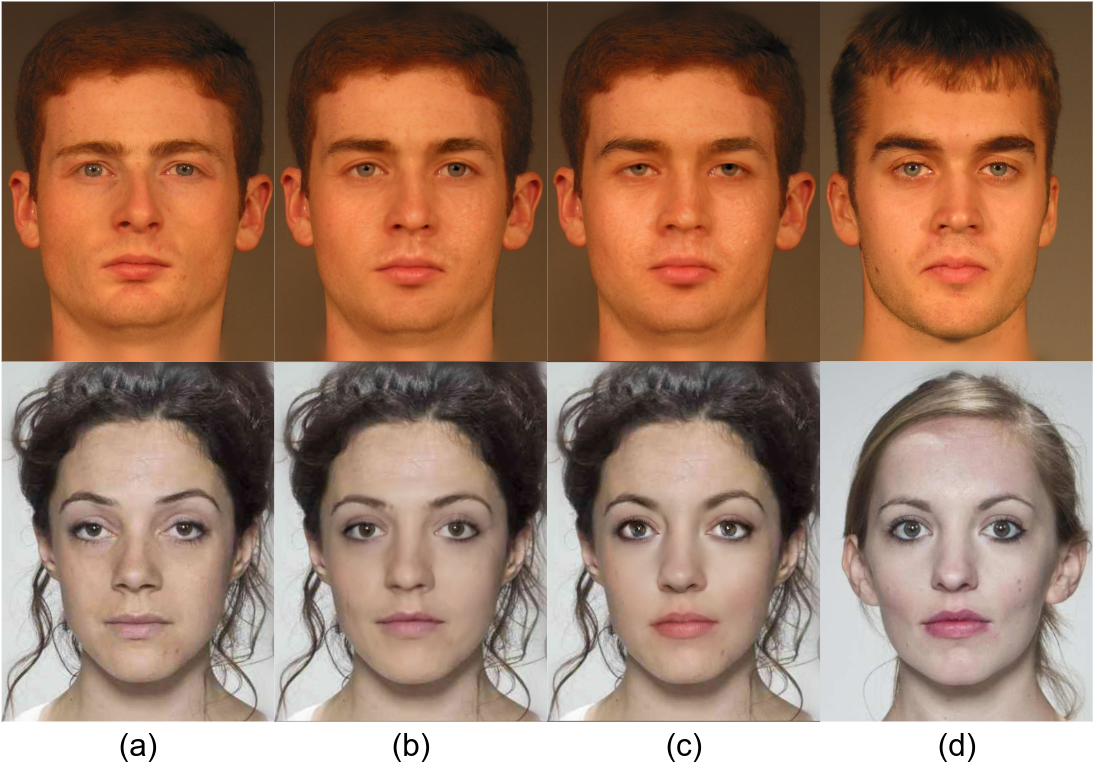}
\end{center}
\caption{Bona fide images (a) and (d) compared to our StyleWarp morphs (b) and morph using PCA norm selection at $p = 10\%$ (c) generated from the FRGCv2 dataset (top row) and the FRLL dataset (bottom row).}
\label{fig::morph_examples}
\end{figure}

We apply our morphing technique on images from the Face Recognition Grand Challenge (FRGCv2) dataset \cite{phillips2005overview} and the Face Research London Lab (FRLL) dataset \cite{debruine2017face}. From the FRGC dataset, we select a subset of pairings used in \cite{zhang2021mipgan} containing 374 bona fide subjects generating 747 morphing pairs to ensure the pairings are comparable to previously published work. We use 102 bona fide subjects from the FRLL dataset and 1140 morphing pairs adapted from \cite{Sarkar2020}. We note that the FRLL images are smaller than $1024\times 1024$. The bona fide subjects are upsampled before finding their latent representations.

\subsection{Training Paradigm}
LPIPS is calculated using the VGG16 \cite{simonyan2014very}. Target and synthesized images are reduced to $256\times 256\times 3$ due to the input size of the VGG16 feature extractor. We apply an Adam optimizer with beta values $\beta_1$ at 0.9 and $\beta_2$ at 0.999 over 1000 iterations to minimize our loss function \ref{total_loss}. We increase the learning rate linearly from 0 to 0.1 for the first 50 steps of optimization and decrease the learning rate using a cosine schedule over the last 600 steps \cite{karras2020analyzing}. The ramp-down duration was increased from 250 steps to 400 due to early convergence during optimization. Avoiding over-fitting and finding local optima are addressed using latent noise and FaceNet verification. Gaussian noise, $N(0,1)$, is applied to the latent code during the first 250 steps at a reducing rate for the first 750 steps to improve latent exploration. We set $T_s$ equal to 400, removing the noise input's influence on the latent code for the remaining 600 steps. Pixel-wise loss leads to the synthesized image becoming smooth; we weight pixel-wise loss by a factor of $0.05$ to prevent smoothing over the output image. Landmark enforcement loss only enforces the landmarks of the synthesized image; it is weighted by a factor of $10^{-4}$ to increase the influence of perceptual and pixel-wise loss over the landmark enforcement. $B$ is initialized with a Gaussian $N(0,1)$. In our total synthesis loss (Equation \ref{total_loss}), we set $\lambda_1$ to $0.05$, $\lambda_2$ to $10^5$, $\lambda_3$ to $0.1$, and $\lambda_4$ to $10^{-4}$.

\begin{table}
\centering
\footnotesize
\captionsetup{justification=centering}
\caption{Single detector performance (left) and MMPMR\% (right) on FRGCv2 dataset.}
\setlength{\tabcolsep}{3.3pt}
\begin{tabular}{|l||cccc||c|}

\hline

    \multirow{2}{*}{Method} & \multicolumn{3}{c}{APCER @ BPCER} & \multicolumn{1}{c||}{EER}&MMPMR \\
        & 1\% & 5\% & 10\% & \%&\% \\
    \hline
    Landmark ~\cite{quek2019face} & 36.46 & 19.84 & 12.33 & 11.17 & 91.49\\
    \hline
    MIPGAN ~\cite{zhang2021mipgan} & 55.19 & 41.56 & 29.22 & 17.54 & 78.00\\
    \hline\hline
    Our StyleWarp    & 91.86 & 72.96 & 64.83 & 33.56 & 79.85      \\
    \hline\hline
    PCA Max $p=80\%$ & 93.09 & 70.33 & 54.73 & 26.60 & 79.79 \\
    \hline
    PCA Max $p=70\%$ & 94.15 & 69.95 & 51.06 & 26.61 & 80.59\\
    \hline
    PCA Max $p=60\%$ & 97.56 & 72.86 & 56.10 & 27.66 & 79.65 \\
    \hline
    PCA Max $p=50\%$ & 94.83 & 69.83 & 59.48 & 28.16 & 79.12\\
    \hline
    PCA Max $p=40\%$ & 96.50 & 65.97 & 48.72 & 26.13 & 78.85\\
    \hline
    PCA Max $p=30\%$ & 83.46 & 59.03 & 44.78 & 24.41 & 78.25\\
    \hline\hline
    PCA Norm $p=80\%$ & 89.24 & 64.26 & 46.39 & 25.95 & 80.66\\
    \hline
    PCA Norm $p=70\%$ & 83.24 & 68.11 & 43.24 & 25.23 & 80.79\\
    \hline
    PCA Norm $p=60\%$ & 91.18 & 78.24 & 66.12 & 30.56 & 79.80\\
    \hline
    PCA Norm $p=50\%$ & 93.54 & 78.46 & 66.77 & 26.77 & 78.92\\
    \hline
    PCA Norm $p=40\%$ & 97.16 & 76.99 & 65.34 & 30.94 & 77.28\\
    \hline
    PCA Norm $p=30\%$ & 95.12 & 74.80 & 60.70 & 28.69 & 74.36\\
    \hline

\end{tabular}
\label{tab::frgc_single}
\end{table}

\subsection{PCA Training}

Our PCA model is trained using style vectors of a self-procured dataset of convex hulls. We organize the dataset of styles by vectors totally 26 matrices of style vectors. Each matrix is used to train a separate PCA model for each style vector. Training a unique model for each style vector is essential due to the varying dimensions of the style vectors. Additionally, each style vector contains different information pertaining to the bona fide image; we desire to morph individual styles and not uncorrelated sets of styles. In Figure \ref{fig::pca_eigen} we show the effect of style-based PCA decomposition by projecting sets of styles onto a varying amount of eigenvectors while projecting the other styles onto a fixed number of eigenvectors. The styles representing coarser information (top) quickly restore the structural features of the original image whereas the styles representing finer information (bottom) quickly restore correct skin tone.

\vspace{-1mm}
\subsection{Results}
We first evaluate our morphs on a single-morph detector to evaluate their performance as a stand-alone image compared to published morphing methods. The detector is a pre-trained FaceNet \cite{schroff2015facenet} model with an additional fully connected layer appended to the end. We train the fully connected layer to classify the input as a real or morph image. The detector is trained on morphs from both landmark-based and StyleGAN2-based techniques using a self-procured dataset. We compare the performance of landmark-based morphs \cite{quek2019face,neubert2018extended}, alternative GAN-based morphs \cite{zhang2021mipgan, Sarkar2020}, our latent averaging morphs (StyleWarp), and morphs using PCA at varying thresholds. We apply these techniques on the FRGCv2 dataset \cite{phillips2005overview} (Table \ref{tab::frgc_single}) and the FRLL dataset \cite{debruine2017face} (Table \ref{tab::frll_single}). Vulnerability analysis is conducted on differential FaceNet verifier \cite{schroff2015facenet} using Mated Morph Presentation Match Rate (MMPMR) \cite{scherhag2017biometric}. MMPMR is computed by comparing the similarity score of each morph to an image of both contributing bona fide subjects. The minimum similarity scores are compared to a fixed threshold to classify each attack as successful or unsuccessful. We use a False Acceptance Rate (FAR) of $10^{-3}$.

\begin{table}[t]
\centering
\footnotesize
\captionsetup{justification=centering}
\caption{Single detector performance (left) and MMPMR\% (right) on FRLL dataset.}
\setlength{\tabcolsep}{3.3pt}
\begin{tabular}{|l||cccc||c|}

\hline

    \multirow{2}{*}{Method} & \multicolumn{3}{c}{APCER @ BPCER} & \multicolumn{1}{c||}{EER}& MMPMR \\
        & 1\% & 5\% & 10\% & \% &\%\\
    \hline
    Landmark ~\cite{neubert2018extended} & 22.54 & 10.78 & 2.94 & 6.34 & 80.13\\
    \hline
    StyleGAN2 ~\cite{Sarkar2020} & 19.49 & 5.93 & 2.54 & 4.98 & 16.12\\
    \hline\hline
    Our StyleWarp & 38.98 & 27.12 & 12.71 & 9.71 & 53.16\\
    \hline\hline
    PCA Max $p=80\%$ & 43.75 & 20.83 & 13.54 & 10.11  & 53.58\\
    \hline
    PCA Max $p=70\%$ & 32.95 & 18.18 & 4.55 & 8.57 & 53.17\\
    \hline
    PCA Max $p=60\%$ & 45.95 & 6.76 & 1.35 & 5.62 & 52.53\\
    \hline
    PCA Max $p=50\%$ & 38.89 & 9.26 & 1.85 & 6.87 & 51.92\\
    \hline
    PCA Max $p=40\%$ & 45.68 & 19.75 & 2.47 & 8.44 & 50.67\\
    \hline
    PCA Max $p=30\%$ & 20.87 & 7.83 & 3.48 & 6.99 & 48.33\\
    \hline\hline
    PCA Norm $p=80\%$ & 27.16 & 22.22 & 7.41 & 7.37 & 54.00\\
    \hline
    PCA Norm $p=70\%$ & 27.93 & 12.61 & 5.41 & 7.14 & 53.67\\
    \hline
    PCA Norm $p=60\%$ & 39.78 & 26.88 & 17.20 & 11.91 & 53.42\\
    \hline
    PCA Norm $p=50\%$ & 42.59 & 25.00 & 13.89 & 10.05 & 51.17\\
    \hline
    PCA Norm $p=40\%$ & 54.87 & 19.51 & 3.66 & 7.30 & 50.42\\
    \hline
    PCA Norm $p=30\%$ & 61.25 & 20.00 & 8.75 & 8.59 & 45.08\\
    \hline

\end{tabular}
\label{tab::frll_single}
\end{table}

Performance of our morphs on the single-morph detector shows the dissimilarity between our morphing approach and both landmark-based and GAN-based morphing techniques. The hybrid approach of our technique produces morphs that the detector is unable to identify as successfully compared to landmark and other StyleGAN-based methods. This performance increase is related to the increased image quality of our morphs compared to the other methods. Performance reduction between the FRGCv2 and the FRLL datasets is related to the pairings used to morph. MMPMR results from our StyleWarp method perform better than both alternative StyleGAN-based morphing methods \cite{zhang2021mipgan, Sarkar2020} with the PCA methods improving upon those results. The vector-wise norm method, however, can result in the morph becoming biased toward one subject if the majority of their projected style vectors have a larger $L_2$ norm as observed by the decrease in MMPMR as $p$ decreases. While the MMPMR results between the GAN-based and landmark-based methods is still prevalent, our StyleWarp and PCA morphs reduce this performance difference.

\begin{figure}[t]
\begin{center}
\includegraphics[width=.92\linewidth]{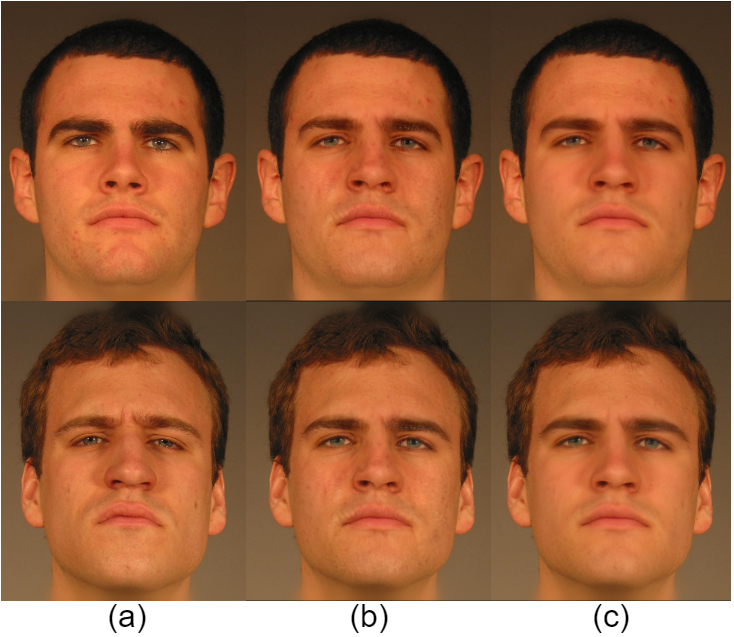}
\end{center}
\caption{Bona fide images (a) compared to morphs using random noise (b) and trained noise (c).}
\label{fig::psnr_morphs}
\end{figure}

\subsection{Noise Trainability}

To further improve our morphs, we explore the noise input of the StyleGAN2 \cite{karras2020analyzing} model. We observe a difference in the facial textures of the bona fide and morphed images. In place of inputting random Gaussian noise into the model, we train for noise values to complement the morphed image. The latent codes of a pair of warped bona fide subjects are averaged and frozen while training for the optimal noise input. The noise values can lead to similar artifacts found in landmark-based morphed images \cite{quek2019face}. For our loss, we calculate the peak signal-to-noise ratio (PSNR) between both bona fide images and the morph and scale the noise values by their root-mean-square after each optimization step. This reduces the number of artifacts present in the morph. In addition, the identity-bias problem is addressed using a scalar for the PSNR values of the contributing subjects using the FaceNet \cite{schroff2015facenet} distance between the morph and the contributing subjects. Our loss function is defined as:
\vspace{-1mm}
\begin{equation}
\small
    \label{psnr}
    L_{psnr} = 20\lambda_5 \log_{10} \frac{255}{||t_1 - g_m||_2} - 20\lambda_6 \log_{10} \frac{255}{||t_2 - g_m||_2},
\end{equation}
where $t_1$ and $t_2$ are the bona fide images, $g_m$ is the synthesized morph image, and $\lambda_5$ and $\lambda_6$ are the identity balance scalars. We apply our algorithm over 200 steps using an Adam optimizer with beta values $\beta_1$ at 0.9 and $\beta_2$ at 0.999 using the FRGCv2 pairings (examples shown in Figure \ref{fig::psnr_morphs}. The performance of these morphs against the single-morph detector are reduced compared to our StyleWarp method; however, performance against FRS verification improves, with a MMMPR of 88.69\%. Noise training increases the threat GAN-based morphs pose to FRS at the cost of increase single-morph detectability.

\vspace{-1mm}
\section{Conclusion}
Our novel morphing method increases the threat of GAN-based morph generation by enforcing the geometric identity and improving the blending of latent representations. The enforcement of landmarks in the image domain improves the performance of GAN-based morphs while masking removes artifacts generation on the outer edges of the images. By limiting noise during training, we improve the calculation of latent representations of the warped convex hulls to increase our morphs' performance. We replace latent averaging with two alternatives using PCA to address identity loss in the morphs. Our method increases the threat of GAN-based morphing to FRS and morph detectors.

\begin{center}
ACKNOWLEDGEMENT
\end{center}
This work is based upon a work supported by the Center for Identification Technology Research and the National Science Foundation under Grant $\#1650474$.

{\small
\bibliographystyle{ieee}
\bibliography{egbib}
}

\end{document}